\documentclass[a4paper, 10 pt, conference]{ieeeconf}

\IEEEoverridecommandlockouts  
\usepackage{pifont}
\usepackage{amsmath,amsfonts}
\usepackage{booktabs}
\usepackage{algorithm}
\usepackage{algpseudocode}
\usepackage{array}
\usepackage{textcomp}
\usepackage{stfloats}
\usepackage{url}
\usepackage{array,multirow,graphicx}
\usepackage{subcaption}
\usepackage{tikz}
\usetikzlibrary{arrows.meta,positioning}
\usepackage{float}
\usepackage{gensymb}
\usepackage{caption}
\usepackage{placeins}
\usepackage{multirow}
\usepackage{flushend}
\usepackage{verbatim}
\usepackage{graphicx}
\usepackage{cite}
\usepackage{hyperref}
\graphicspath{{./Fig/}}	
\usepackage[english]{babel}

\DeclareUnicodeCharacter{2061}{}
\makeatletter
\renewcommand{\fnum@figure}{Fig. \thefigure}
\makeatother
\renewcommand{\baselinestretch}{0.98}
\usepackage[flushleft]{threeparttable}

\begin{document}
\title{\LARGE \bf Vortex-Induced Torque in Grasping Soft Objects: \\   Influence of Geometry and Pneumatic Parameters}

\author{Roman Mykhailyshyn$^{1}$, Tomohiro Motoda$^{2}$, Ivan Virgala$^{3}$, Ann~Majewicz Fey$^{4}$,  and Yukiyasu Domae$^{5}$
\thanks{$^{1}$Roman Mykhailyshyn is with the National Institute of Advanced Industrial Science and Technology (AIST), Tokyo 135-0064, Japan. (Corresponding author e-mail: {roman.mykhailyshyn@aist.go.jp})}%
\thanks{$^{2}$Tomohiro Motoda is with the National Institute of Advanced Industrial Science and Technology (AIST), Tokyo 135-0064, Japan.  e-mail: tomohiro.motoda@aist.go.jp}%
\thanks{$^{3}$Ivan Virgala is with the  Department of Industrial Automation and Mechatronics, Technical University of Košice, 04200 Košice, Slovak Republic. {ivan.virgala@tuke.sk}}%
\thanks{$^{4}$Ann Majewicz Fey is with the Department of Mechanical Engineering, The University of Texas at Austin, Austin, TX 78712, USA. {Ann.MajewiczFey@utexas.edu}}%
\thanks{$^{5}$Yukiyasu Domae is with the Embodied AI Research Team, National Institute of Advanced Industrial Science and Technology (AIST), Tokyo 135-0064, Japan. e-mail: domae.yukiyasu@aist.go.jp}%
}



\maketitle
\thispagestyle{empty}
\pagestyle{empty}

\begin{abstract}
Vortex grippers are commonly integrated into automated systems for handling objects with complex geometries in packaging, sorting, and assembling. However, their interaction with compliant objects remains difficult to predict, particularly with respect to the vortex-induced torque transmitted during grasping. This study investigates vortex-induced torque during the grasping of compliant objects and quantifies the coupled effects of supply pressure, tangential nozzle diameter, surface geometry, and curvature. The results show that increasing the tangential nozzle diameter primarily enhances the sensitivity of the generated torque to supply pressure rather than producing a uniformly higher torque. Surface geometry also strongly influences the rotational interaction, with dome-concave surfaces generally producing substantially greater torque than corresponding dome-convex surfaces, whereas the difference between cylindrical concave and convex surfaces is comparatively small. The strongest and most repeatable response was observed for a dome-concave surface exhibiting favorable object-gripper geometric matching. Among the evaluated machine-learning regression methods, Gaussian Process Regression provided the best overall predictive performance under repeated cross-validation, demonstrating that vortex-induced torque can be estimated from combined pneumatic and geometric parameters. These findings provide a basis for predicting and utilizing vortex-induced torque in the dexterous manipulation of soft objects.
\end{abstract}

\section{Introduction}

Pneumatic gripping devices are very easy to integrate into robotic automation processes due to their predictability and ease of use~\cite{monkman2007robot, 9721534}. This is a general statement that is only partially true. There are many different subtypes of pneumatic gripping devices~\cite{mykhailyshyn2022systematic, papadopoulos2023deformable}: vacuum~\cite{6630601, schaffrath2021vacuum, MANTRIOTA200718} (pumpless, piston, ejector, pumping); air-jet~\cite{park2012design, machines11060667, li2011computational, Mykhailyshyn2022gripper} (vortex, Bernoulli, combined); combined~\cite{Fujita19032018, michalos2018novel, derby2005robotic, tawk20193d} (pneumatic-pneumatic, pneumatic-other). The complexity of their integration into automated systems is that most of them are built on different physical principles. This does not easily allow for predicting the behavior of the gripper with different objects, especially soft ones.

\begin{figure*}[htbp]
\newcommand{\mywidth}{1}
\centering
	\begin{subfigure}[b]{0.23\linewidth}
         \centering
         \vspace{2mm}
         \includegraphics[width=1\linewidth, clip, trim=0pt 0pt 0pt 0pt]{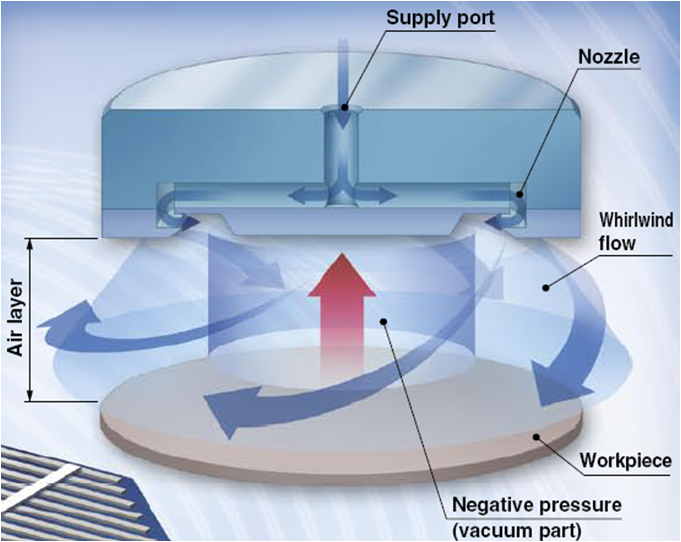}
         \vspace{-5.5mm}
         \caption{}
         \label{subfig2_1}
	\end{subfigure}%
    ~
	\begin{subfigure}[b]{0.253\linewidth}
         \centering
         \vspace{2mm}
         \includegraphics[width=1\linewidth, clip, trim=0pt 0pt 0pt 0pt]{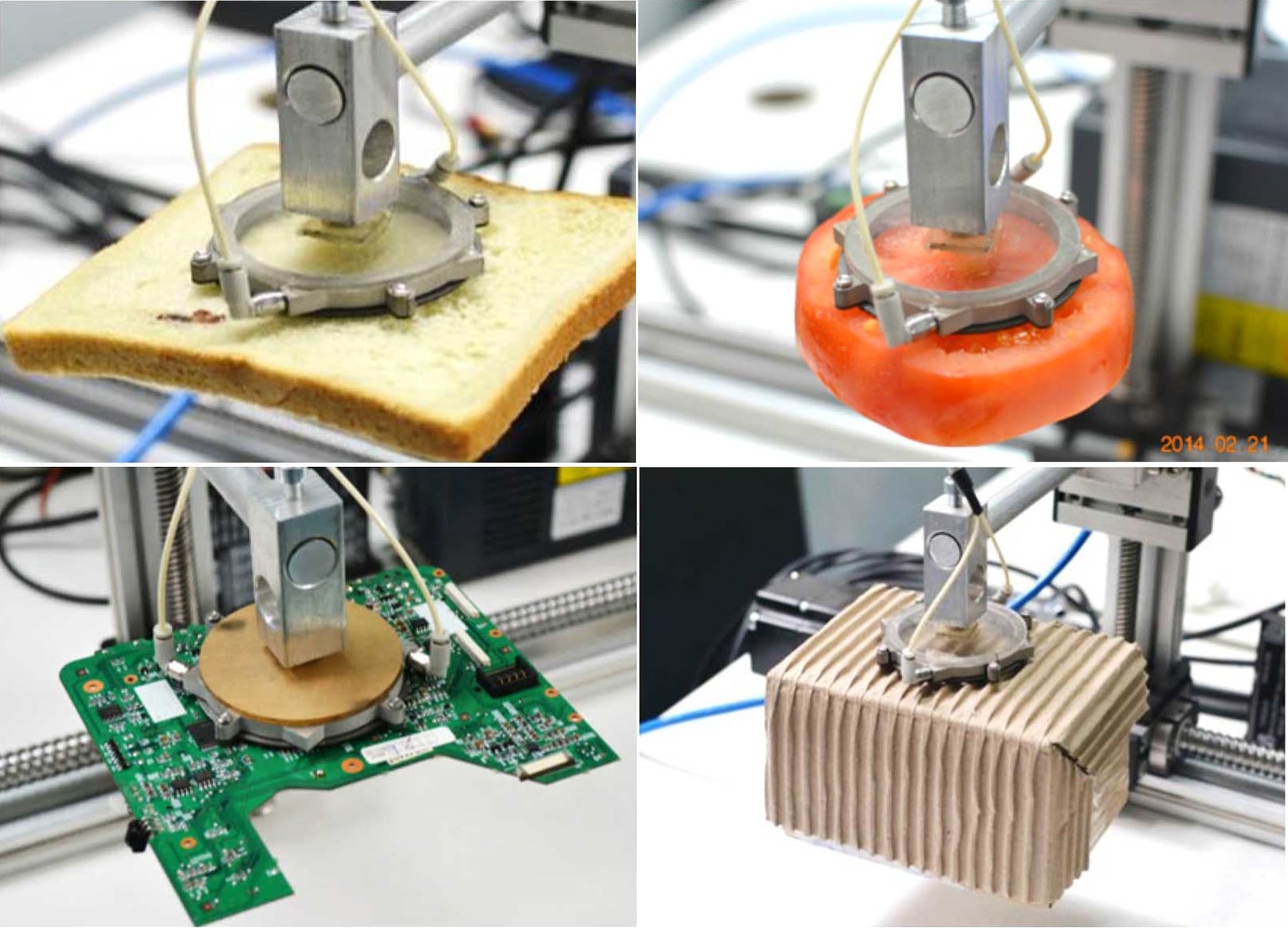}
         \vspace{-5.5mm}
         \caption{}
         \label{subfig2_2}
	\end{subfigure}%
    ~
	\begin{subfigure}[b]{0.279\linewidth}
         \centering
         \vspace{2mm}
         \includegraphics[width=1\linewidth, clip, trim=0pt 0pt 0pt 0pt]{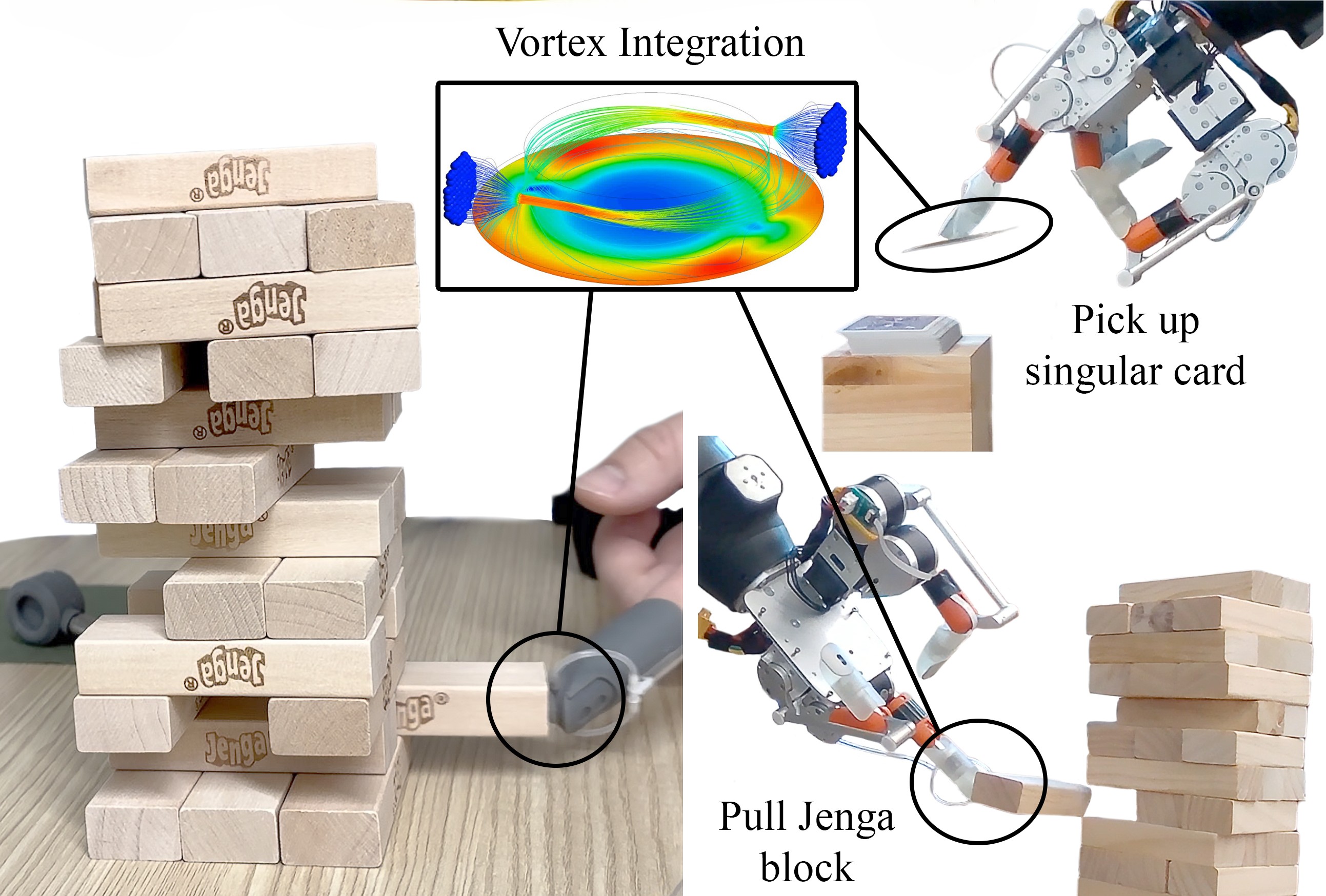}
         \vspace{-5.5mm}
         \caption{}
         \label{subfig2_3}
	\end{subfigure}%
    ~
	\begin{subfigure}[b]{0.21\linewidth}
         \centering
         \vspace{2mm}
         \includegraphics[width=1\linewidth, clip, trim=0pt 0pt 0pt 0pt]{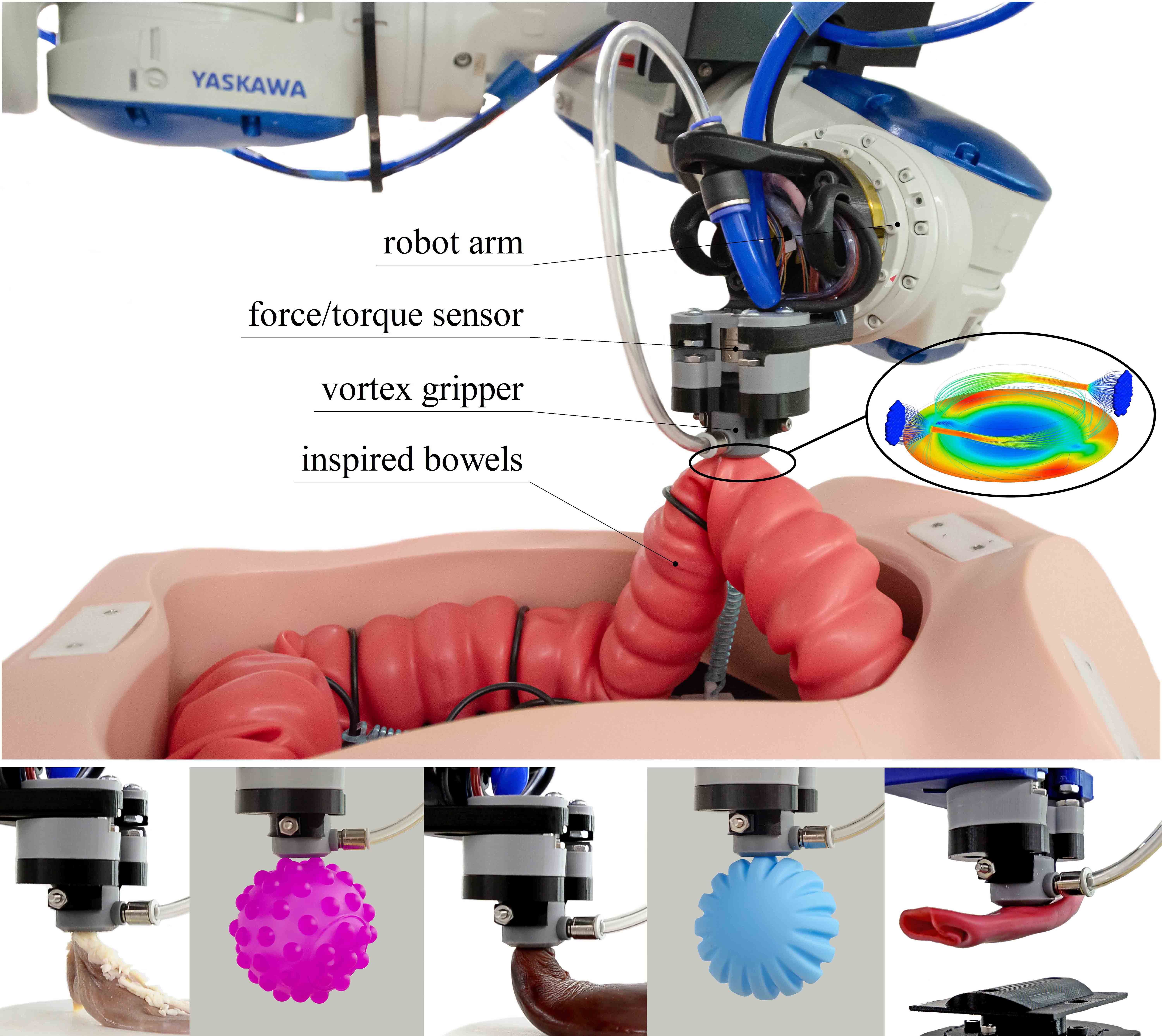}
         \vspace{-5.5mm}
         \caption{}
         \label{subfig2_4}
	\end{subfigure}%
    \vspace{-1.5mm}
\caption{Vortex gripper technology and representative applications: (a) compressed air flow direction that form a vortex rotation flow that creates a lifting effect~\cite{mykhailyshyn2022systematic}; (b) application of vortex technology for grasping objects in manufacturing and assembly (electronics, food, packaging, etc.)~\cite{RoughSurfaceSucker}; (c) integration of soft vortex gripper into the dexterous hand finger~\cite{mykhailyshyn2026icra}; (d) vortex gripper for grasping soft tissue of various shapes~\cite{mykhailyshyn2024low, roman2026investigating}.}
\label{fig2}
\end{figure*}

By considering the most widely used suction cups based on ejector systems, it is obvious that the behavior of the object can be easily predicted by uniform distribution of air pressure by the suction cup over the surface of the object~\cite{11593517, sentis2026}. However, when it comes to air-jet grippers, the interaction between the gripper and the object becomes very complex~\cite{li2011computational}. This can be seen if we do not consider soft objects but only rigid objects; changing the parameters of the gripper geometry~\cite{machines11060667, chandran2019numerical}, nozzle~\cite{lyu2025effect}, object geometry~\cite{shi2024development}, supply pressure~\cite{liu2018soft, zhao2022study}, and object-gripper gap~\cite{wang2019effect, zhao2019gap, renn2008gap} leads to significant changes in both the object deformation and the force characteristics of the gripper. This is due to the fact that the pressure distribution on the surface of the object will not be uniform (Fig.~\ref{subfig2_1}), and zones of negative and positive pressure will be formed~\cite{ozcelik2005examination, Wang_2025}.

Such characteristics of jet grippers have both advantages and disadvantages; for example, they allow grasping an object from a distance~\cite{mykhailyshyn2022influence, SHI2018323, seita2021learning}, but they create additional vibrations~\cite{mykhailyshyn2025vibration}. Due to the ability of vortex and Bernoulli grippers to hold objects that are not completely connected to the gripper (like suction cups failing under these conditions), they have gained popularity for grasping fragile~\cite{mykhailyshyn2022three}, porous~\cite{Mykhailyshyn2022gripper}, or electronic parts~\cite{RoughSurfaceSucker, SUN2026114979} (Fig.~\ref{subfig2_2}). Additionally, the ability to grasp objects from a distance~\cite{cao2025noncontact} has expanded the scope of applications of jet grippers for deformable~\cite{mykhailyshyn2026sii, 11488526, hanamura2024fabric} and soft~\cite{10937130} objects. They are most widely used for grasping textile objects from a stack~\cite{liu2020design, Mykhailyshyn2022gripper} and for local control of the position of deformable materials in manufacturing~\cite{mykhailyshyn2023fem, 11037411}. Many studies have focused on the use of jet grippers for both dexterous manipulation through their integration into the fingers~\cite{VacuumFingertips, mykhailyshyn2026icra} of anthropomorphic hands (Fig.~\ref{subfig2_3}) and for dexterously manipulating deformable objects~\cite{Mykhailyshyn2024toward} with the robot arm.

However, unlike Bernoulli grippers, vortex grippers have additional characteristics that positively affect the lifting force due to a greater ability to adapt to the surface of the object~\cite{li2015experimental}, as well as negatively, which leads to the rotation of the object during grasping. This is because the lifting force is created by the rotation of the airflow, forming a vortex that lifts the object but also creates a torque that rotates the object around the axis of the gripper (Fig.~\ref{subfig2_1}). Researchers and developers of vortex grippers usually avoid discussing this negative factor and solve the problem by integrating friction elements, which also works for other jet grippers~\cite{mykhailyshyn2021substantiation}. Also, if it is not possible to use friction elements, side stops are integrated to prevent the object from rotating. By using wheels as friction elements, we can see the application of vortex technology to mobile robots moving along walls~\cite{zhao2018experimental}.


Despite the torque effect, vortex grippers have many positive effects that allow them to be used for soft tissue grasping in medical applications. In our previous studies~\cite{mykhailyshyn2024low, roman2026investigating},
we observed the rotation of bio-inspired tissue during vortex
grasping and mitigated this effect using friction elements
(Fig.~\ref{subfig2_4}). However, the influence of compliant-object geometry on vortex-induced torque remains insufficiently characterized, limiting the ability to predict rotational interaction during soft-object grasping. A quantitative understanding of this torque may also enable future manipulation strategies in which the vortex-induced rotational effect is deliberately exploited rather than suppressed. It is also important to understand that knowledge of torque parameters can enable in-hand manipulation without dexterous movements, using only vortex flows. Therefore, in this paper, a detailed study of vortex-induced torque in grasping soft objects was conducted, with 4920 valid trials distributed over 492 experimental conditions. The main contributions of this study are summarized as follows:
\begin{itemize}%
    \item Experimental quantification of vortex-induced torque as a function of supply pressure, tangential nozzle diameter, surface geometry, and curvature was conducted.

    \item Identification of geometry-dependent torque amplification for dome-concave surfaces, including a highly repeatable local maximum near a specific object-gripper geometric relationship.

    \item Demonstration that lifting force and vortex-induced torque exhibit different dependencies on the gripper-surface gap, showing that rotational interaction cannot be inferred from lifting force behavior alone.

    \item Introduction of an object-gripper geometric-matching interpretation to describe the observed torque amplification and the differences between dome and cylindrical surfaces.

    \item Development and cross-validated comparison of machine-learning regression models for predicting peak vortex-induced torque from pneumatic and geometric parameters, with Gaussian Process Regression providing the best overall performance.
\end{itemize}

\section{Methodology} \label{met}

\subsection{Experimental Setup}

\begin{figure}[!t]
    \centering
    \vspace{-2mm}
    \includegraphics[width=1\linewidth,clip ,trim=0pt 0pt 0pt 0pt]{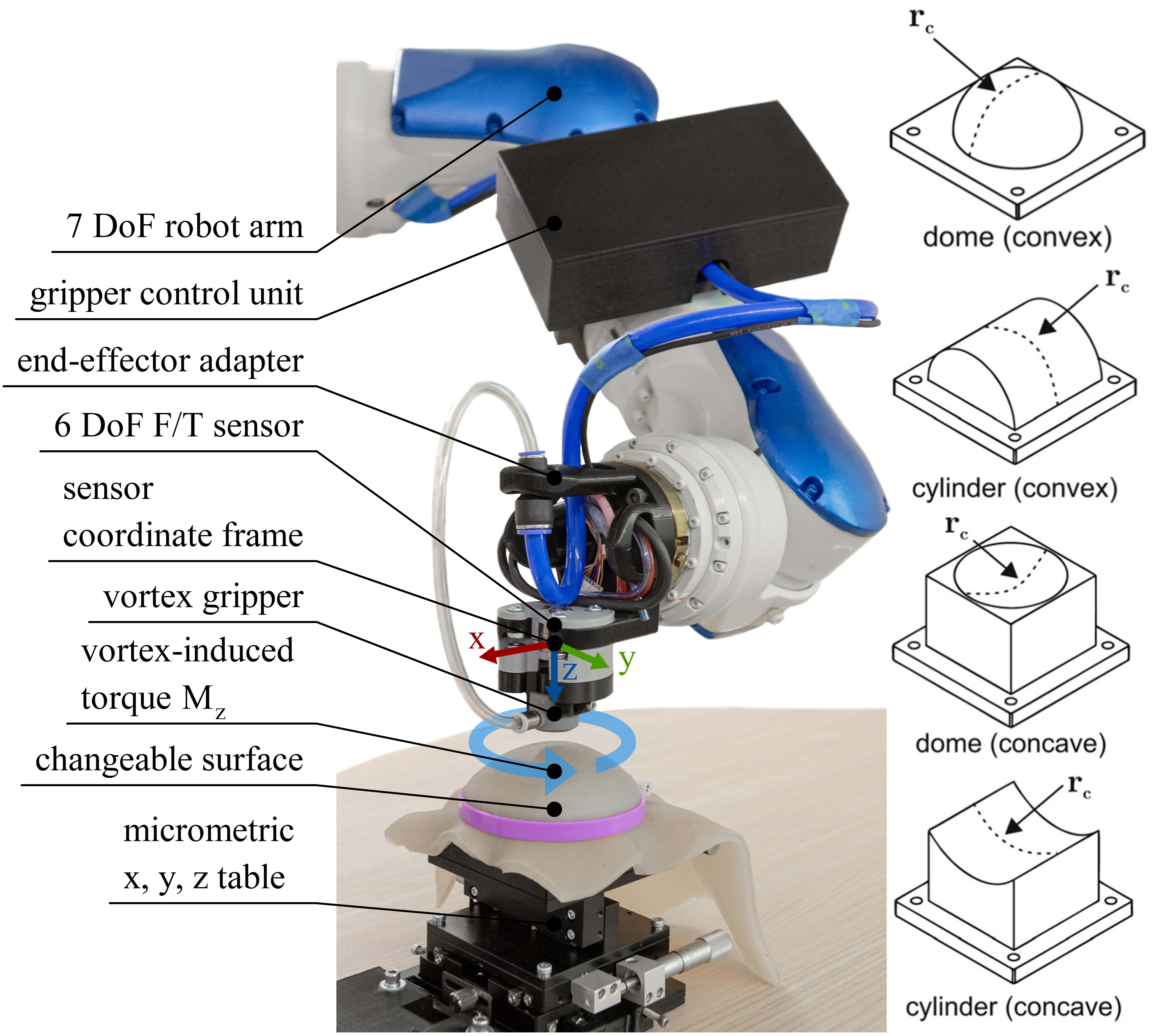}
    \caption{Experimental setup with different object surfaces for studying the force characteristics of a vortex gripper.}
    \label{fig_3}
\end{figure}

The experiments were conducted to quantify the rotational interaction generated by the vortex flow during gripper contact with compliant surfaces and to determine how this interaction is influenced by supply pressure, tangential nozzle diameter, and local surface geometry. The primary response variable was the torque about the vortex axis, $M_z$, whereas the axial force $F_z$ was additionally used to identify the transition from mechanically compressed contact to the vortex lifting stage (Fig.~\ref{fig_3}).

The experimental setup (Fig.~\ref{fig_3}) consists of a 7-DOF robot arm with an attached gripper control unit and an end-effector adapter. A vortex gripper is attached to the end-effector adapter via an ATI Nano17 force/torque sensor with an accuracy of ±0.25$\%$. Three geometrically identical vortex grippers were investigated. The grippers had an outer diameter of $D_g = 20$~mm, a vortex-chamber diameter of $D_c = 14$~mm, and a chamber depth of $H_c = 4$~mm. Each gripper contained two tangential nozzles. The investigated nozzle diameters were $d_n$ in 0.6, 0.8, 1.0~mm for corresponding gripper configurations G$_1$, G$_2$, and G$_3$, respectively. Four supply pressures were investigated, $P_s$ in 100, 200, 300, 400~kPa. The compliant test represented four curved surface types: cylindrical concave, cylindrical convex, dome concave, and dome convex. Ten nominal surface radii were studied, $R$ in 15, 20, 25, 30, 35, 40, 45, 50, 75, 100~mm. A compliant flat surface was additionally investigated. Thus, each gripper configuration was evaluated on 41 different surface geometries. The interaction layer was based on standardized bio-inspired silicone soft tissue, which was fabricated from two-component silicones (Smooth-On) and the two different layers, representing skin and subcutaneous fat~\cite{roman2026investigating}.

During each surface experiment, the gripper was first brought into contact with the soft surface and pressed against it. Compressed air was subsequently supplied while the gripper-surface interface remained initially closed. The gripper was then moved vertically upward at a constant velocity of $v_z = 10$~mm/s, thereby opening the gap between the gripper and the compliant surface. Ten repetitions were performed for every combination of nozzle diameter, pressure, and surface geometry. The complete nominal factorial design therefore contained 4920 individual trials. 

The measured $M_z$ (Fig.~\ref{fig_3}) represents the reaction torque acting on the gripper rather than a direct measurement of the torque transmitted exclusively to the grasped object. The measured signal can additionally contain contributions from the free airflow, structural loading, and other quasi-static effects. For this reason, the largest raw $|M_z|$ value recorded over the complete robot trajectory was not used directly. The force and torque signals were acquired at a sampling frequency of $f_s = 2000$~Hz. The sign convention was selected such that negative $F_z$ represents mechanical compression of the gripper, whereas positive $F_z$ corresponds to the direction of the lifting force. The beginning of the analyzed vortex-induced torque was determined directly from the raw $F_z$ signal, where $i_0$ is the first sample satisfying $F_z(i_0-1)<0$, $F_z(i_0)\geq0$.

\subsection{Statistical Analysis}
\label{subsec:statistics}

For each combination of gripper configuration, pressure, and surface geometry, the mean peak torque was calculated from the $n$ available repetitions as

\begin{equation}
\overline{M}_{z,\mathrm{peak}} = \frac{1}{n} \sum_{j=1}^{n} M_{z,\mathrm{peak}}^{(j)}.
\label{eq:mean_peak_torque}
\end{equation}

Trial-to-trial variability was represented by the sample standard deviation,

\begin{equation}
\sigma_{M_z} = \sqrt{ \frac{1}{n-1} \sum_{j=1}^{n} \left( M_{z,\mathrm{peak}}^{(j)} - \overline{M}_{z,\mathrm{peak}} \right)^2 }.
\label{eq:torque_std}
\end{equation}

The coefficient of variation was additionally calculated as $CV = 100 ( \sigma_{M_z} /  \overline{M}_{z,\mathrm{peak}}) $. To summarize the effect of pressure while reducing distortion from very large percentage changes, a logarithmic pressure-sensitivity metric was defined as

\vspace{-2mm}
\begin{equation}
S_P = \log_2 \left( \frac{
\overline{M}_{z,\mathrm{peak}}(400~\mathrm{kPa}) }{
\overline{M}_{z,\mathrm{peak}}(100~\mathrm{kPa}) } \right).
\label{eq:pressure_log_metric}
\end{equation}

This metric has a direct interpretation:

\vspace{-2mm}
\begin{equation}
S_P=
\begin{cases}
0, & \text{no change},\\
+1, & \text{twofold increase},\\
-1, & \text{twofold decrease}.
\end{cases}
\label{eq:pressure_log_interpretation}
\end{equation}

Similarly, the influence of curvature orientation was quantified by matching the concave and convex surfaces at identical surface shape, radius, pressure, and nozzle diameter:

\begin{equation}
S_{c/v}
=
\log_2
\left(
\frac{
\overline{M}_{z,\mathrm{peak}}^{\mathrm{concave}}
}{
\overline{M}_{z,\mathrm{peak}}^{\mathrm{convex}}
}
\right).
\label{eq:concave_convex_metric}
\end{equation}

Thus, $S_{c/v}>0$ indicates greater torque for the concave geometry, while $S_{c/v}<0$ indicates greater torque for the corresponding convex geometry. Where monotonic radius trends were investigated using the Spearman correlation coefficient:

\vspace{-2mm}
\begin{equation}
\rho_s = \operatorname{corr} \left[
\operatorname{rank}(R),
\operatorname{rank}\!\left(\overline{M}_{z,\mathrm{peak}}\right)
\right].
\label{eq:spearman}
\end{equation}

The final analysis contained 4920 valid trials distributed over 492 experimental conditions. Despite the compliant nature of the specimens and the transient character of the interaction, the measurements exhibited generally good repeatability. The coefficient of variation was below 10\% for 26.0\% of the investigated conditions, below 20\% for 69.3\%, below 30\% for 88.0\%, and below 50\% for 97.4\%.

\subsection{Object-Gripper Relationship}
\label{subsec:geometry_matching_method}

For a spherical surface of radius $R$, the geometric deviation from the tangent plane at radial position $r$ can be expressed as $ \delta(r,R) = R-\sqrt{R^2-r^2}$. Two radial positions are particularly relevant to the present gripper: $r_c = D_c/2 = 7$~mm, $r_g = D_g/2 = 10$~mm, corresponding to the vortex-chamber radius and the outer gripper radius, respectively.

To express the magnitude of the surface curvature relative to the chamber depth, the dimensionless geometric ratios were defined as $\Lambda_c(R) =  \delta(r_c, R) / H_c $, $\Lambda_g(R) = \delta(r_g, R) / H_c$ (Table~\ref{tab:geometry_matching}). For a comparatively small peripheral gap, the available annular outlet area scales approximately as $A_{\mathrm{out}} \sim 2\pi r_g g(r_g)$, where $g(r_g)$ denotes the local peripheral gap at the outer
gripper radius. Consequently, relatively small changes in the peripheral gap can produce substantial changes in the effective airflow outlet area.

\begin{table}[t]
\centering
\scriptsize
\caption{Geometric Deviation of the Ideal Dome Relative to the Gripper Dimensions}
\label{tab:geometry_matching}
\setlength{\tabcolsep}{3.0pt}
\renewcommand{\arraystretch}{1.12}

\resizebox{\columnwidth}{!}{%
\begin{tabular}{|c|cccccccccc|}
\hline
\textbf{$R$ [mm]}
& 15 & 20 & 25 & 30 & 35 & 40 & 45 & 50 & 75 & 100 \\
\hline
\textbf{$\delta(7,R)$ [mm]}
& 1.734 & 1.265 & 1.000 & 0.828 & 0.707 & 0.617 & 0.548 & 0.492 & 0.327 & 0.245 \\

\textbf{$\delta(10,R)$ [mm]}
& 3.820 & 2.679 & 2.087 & 1.716 & 1.459 & 1.270 & 1.125 & 1.010 & 0.670 & 0.501 \\

\textbf{$\Lambda_g$}
& 0.955 & 0.670 & 0.522 & 0.429 & 0.365 & 0.318 & 0.281 & 0.253 & 0.167 & 0.125 \\
\hline
\end{tabular}}
\end{table}

At $R=15$~mm, the geometric deviation at the outer gripper radius is $\delta(10,15)=3.820$~mm, which is close to the complete vortex-chamber depth of 4~mm. The corresponding dimensionless value is $\Lambda_g(15)=0.955$. At $R=20$~mm, $\delta(10,20)=2.679$~mm, and $\Lambda_g(20)=0.670$. Thus, the $R=15$-20~mm surfaces modify the gripper-surface geometry on the same dimensional scale as the vortex chamber itself.

\subsection{ML for Torque Prediction}
\label{subsec:ml_method}

Peak vortex-induced torque was predicted using 492 condition-level means, each obtained from ten repetitions; individual trials were not treated as separate ML samples. The input vector was
\vspace{-2mm}
\begin{equation}
\mathbf{x}=[p,\;A_n,\;\Lambda_c^{\pm},\;N_{\kappa}]^{\mathrm{T}},
\label{eq:ml_input}
\end{equation}
where $p=P_s/(400\,\mathrm{kPa})$, $A_n=\pi d_n^2/2$ is the total nozzle area in $\mathrm{mm}^2$, and $N_{\kappa}=0,1,2$ for flat, cylindrical, and dome surfaces, respectively. For curved surfaces,
\vspace{-2mm}
\begin{equation}
\Lambda_c^{\pm}
=s\frac{R-\sqrt{R^2-r_c^2}}{H_c},
\quad
s=
\begin{cases}
+1,&\text{concave},\\
-1,&\text{convex},
\end{cases}
\label{eq:ml_curvature}
\end{equation}
with $r_c=7$~mm and $H_c=4$~mm; $\Lambda_c^{\pm}=0$ for flat surfaces.

Five models were compared: GPR, Random Forest (RF), Extra Trees, AdaBoost, and an equally weighted RF--AdaBoost ensemble. RF and Extra Trees used 700 trees, unrestricted depth, minimum split/leaf sizes of 2/1, and all features at each split; bootstrap sampling was enabled only for RF. AdaBoost used up to 500 trees of maximum depth 4, minimum leaf size 2, learning rate 0.03, and square loss. These settings were fixed without a hyperparameter search.

GPR used a constant-scaled Mat\'ern-$5/2$ kernel with four feature-specific length scales and additive white noise. Initial values (bounds) were 1 ($10^{-3}$-$10^3$) for the constant, 1 ($10^{-2}$-$10^2$) for each length scale, and 0.02 ($10^{-6}$-$10^1$) for noise variance, with diagonal regularization of $10^{-8}$. Kernel parameters were optimized by maximizing the training-fold log marginal likelihood without optimizer restarts. GPR inputs and targets were standardized using training-fold statistics, and predictions were returned to torque units. Tree-based models used no additional standardization.

Evaluation used repeated five-fold cross-validation with five repetitions, identical splits for all models, and random seed 42 for splitting and stochastic estimators. Nested cross-validation was not used. All fitting and GPR preprocessing were performed within each training fold. The $R^2$, RMSE, and MAE were reported as mean $\pm$ standard deviation across 25 validation folds; this standard deviation is not a confidence interval. The model with the highest mean $R^2$ was selected.

\section{Results and Discussion}

\subsection{Rigid Surface Quasi-Static Interaction}
\label{subsec:rigid_results}

\begin{figure}[t]
\centering
\includegraphics[width=0.88\columnwidth]{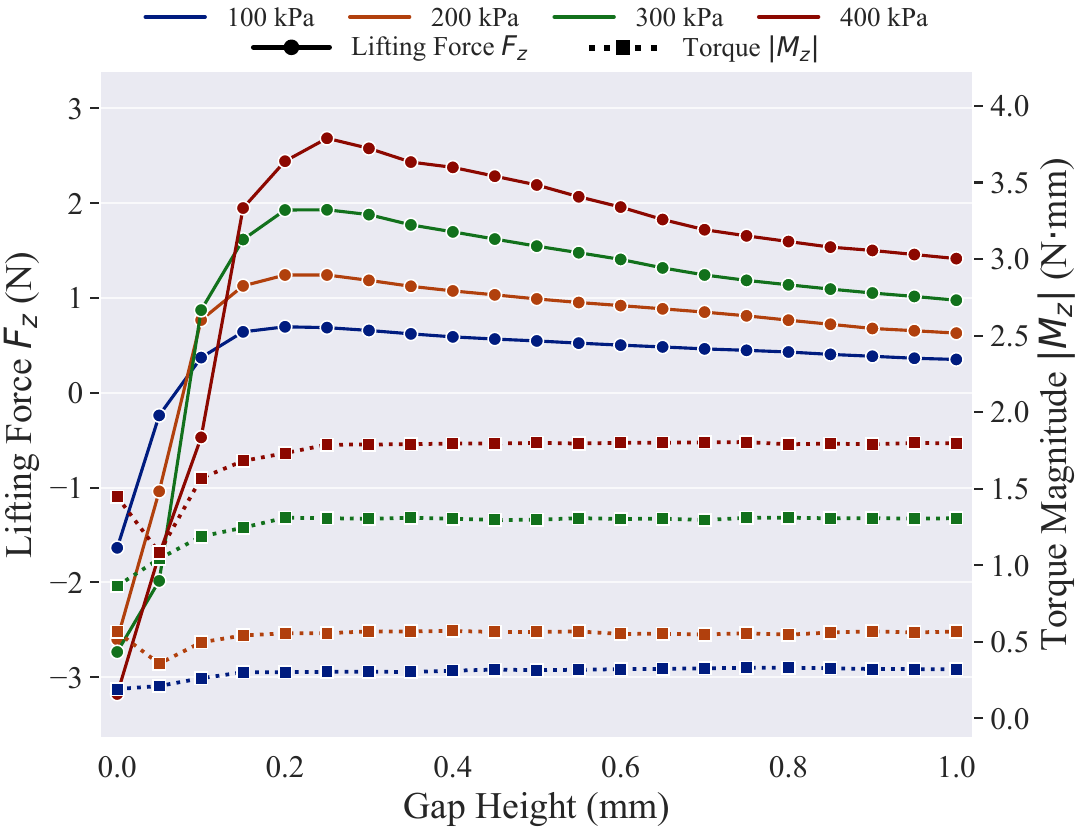}
\vspace{-1.5mm}
\caption{Quasi-static lifting force $F_z$ and torque magnitude $|M_z|$ as functions of gap height for the rigid flat surface using G$_1$.}
\label{fig:rigid_combined}
\end{figure}

A rigid-surface reference experiment was conducted to independently
characterize the dependence of lifting force and vortex-induced torque on the gripper-surface gap when surface deformation was eliminated (Fig.~\ref{fig:rigid_combined}). For this, the G$_1$ gripper with 0.6 mm tangential nozzles and a rigid flat surface was used. The gripper was positioned at discrete stationary gap heights $h =$ 0, 0.05, 0.10,...,1.00~mm. At every height, 100 force/torque sensor readings were taken, and their mean was recorded as the response at the corresponding gap height. The procedure was repeated for supply pressures of 100, 200, 300, and 400~kPa. The transition from mechanically compressed contact to positive lifting force was characterized by the gap height $h_0$ at which $F_z(h_0)=0$. For adjacent measured heights $h_k$ and $h_{k+1}$ satisfying $F_z(h_k)<0$, $F_z(h_{k+1})\geq0$ the zero-crossing height was estimated using linear interpolation:

\begin{equation}
h_0 = h_k - F_z(h_k) \frac{h_{k+1}-h_k}{F_z(h_{k+1})-F_z(h_k)}.
\label{eq:rigid_zero_interpolation}
\end{equation}

The maximum lifting force was defined as $F_{z,\max}^{\mathrm{rigid}} = \max_h F_z(h)$, and the magnitude of the rotational response was calculated as $M_{z,\mathrm{rigid}}(h) = |M_z(h)|$. The corresponding mean torque was calculated as

\vspace{-2mm}
\begin{equation}
\overline{M}_{z,\mathrm{rigid}} = \frac{1}{N_h} \sum_{q=1}^{N_h} |M_z(h_q)|.
\label{eq:rigid_plateau_mean}
\end{equation}

The transition from mechanical compression to positive lifting force occurs at progressively greater gaps as the supply pressure increases. The interpolated zero-crossing heights are 0.070, 0.079, 0.085, 0.110~mm for 100, 200, 300, and 400~kPa, respectively. The maximum occurs at $h=0.20$~mm for 100~kPa and at $h=0.25$~mm for 200, 300, and 400~kPa.

\begin{table}[t]
\centering
\scriptsize
\caption{Quasi-Static Rigid-Flat Surface Interaction}
\label{tab:rigid_results}
\setlength{\tabcolsep}{2.5pt}
\renewcommand{\arraystretch}{1.15}
\begin{tabular}{|c|c|c|c|c|}
\hline
\textbf{$P_s$ [kPa]} &
\textbf{$h_0$ [mm]} &
\textbf{$F_{z,\max}$ [N]} &
\textbf{$h(F_{z,\max})$ [mm]} &
\textbf{$|\overline{M_z}|$ [N$\cdot$mm]} \\
\hline
100 & 0.070 & 0.696 & 0.20 & $0.319\pm0.008$ \\
\hline
200 & 0.079 & 1.243 & 0.25 & $0.560\pm0.008$ \\
\hline
300 & 0.085 & 1.931 & 0.25 & $1.304\pm0.005$ \\
\hline
400 & 0.110 & 2.685 & 0.25 & $1.796\pm0.005$ \\
\hline
\end{tabular}
\end{table}

In Table~\ref{tab:rigid_results}, $|\overline{M_z}|$ represents the mean torque magnitude over the developed-gap interval $0.30\leq h\leq1.00$~mm. The maximum lifting force increases from 0.696~N at 100~kPa to 2.685~N at 400~kPa. Over the four investigated pressure levels, a linear regression gives $F_{z,\max}^{\mathrm{rigid}} \approx 0.00665P_s-0.0248$, with $R^2=0.995$. This experiment is used only as a compact description of the measured pressure range and should not be interpreted as a general physical model of the vortex gripper. The torque shows a substantially different dependence on gap height. Once the gap exceeds approximately 0.30~mm, the torque magnitude becomes nearly independent of further increases in height. The fitted slopes over $0.30\leq h\leq1.00$~mm are only 0.028, -0.011, 0.006, 0.005~N$\cdot$mm/mm at 100, 200, 300, and 400~kPa, respectively.

Consequently, $F_z(h)$ and $M_z(h)$ show fundamentally different gap dependencies. The lifting force has a distinct optimum gap near 0.20-0.25~mm and subsequently decreases with increasing gap. In contrast, once the vortex provides a lifting force, the rotational response remains approximately constant over a considerably wider gap interval.

\subsection{Impact on Torque Sensitivity}
\label{subsec:flat_results}

\begin{figure}[t]
\centering
\begin{subfigure}[t]{0.48\linewidth}
\centering
\includegraphics[width=\linewidth]{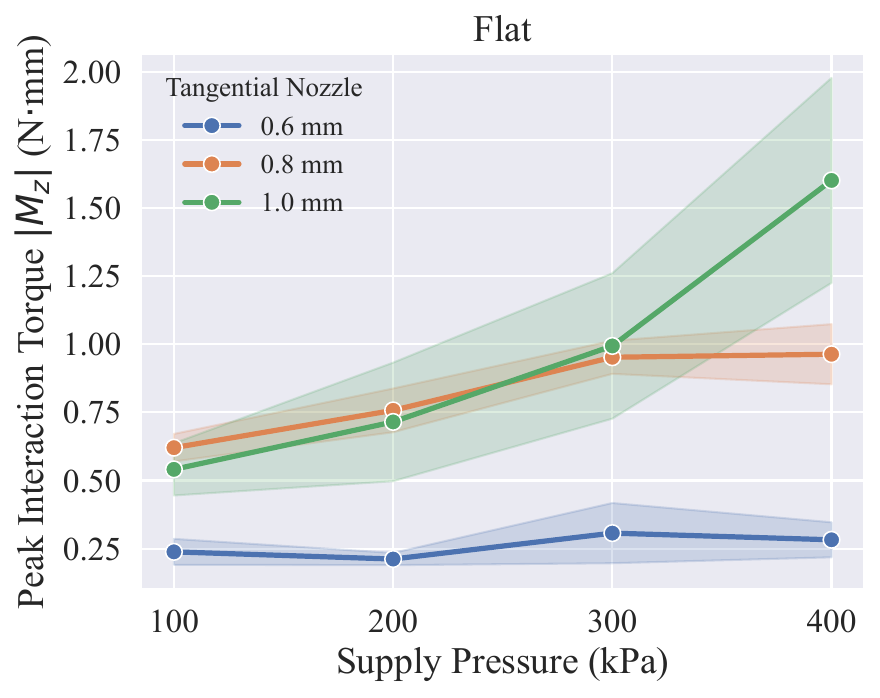}
\caption{}
\label{fig:flat_pressure}
\end{subfigure}
\hfill
\begin{subfigure}[t]{0.48\linewidth}
\centering
\includegraphics[width=\linewidth]{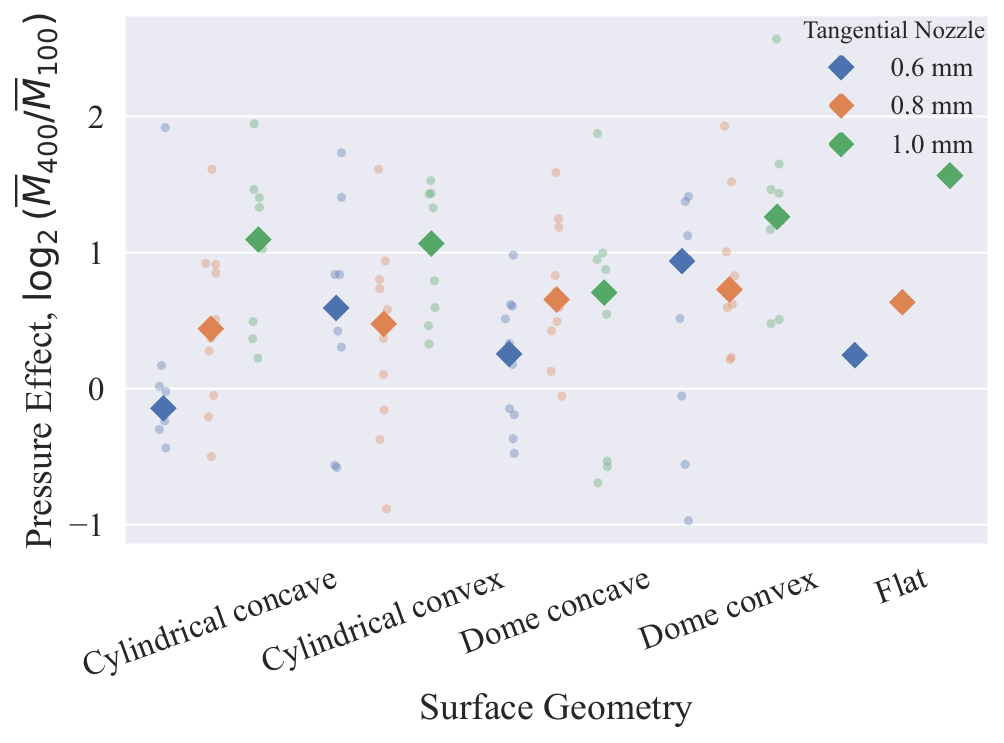}
\caption{}
\label{fig:pressure_log}
\end{subfigure}
\vspace{-2mm}
\caption{(a) Vortex-induced torque for the flat compliant surface as a function of supply pressure. (b) Logarithmic pressure sensitivity of the torque between 100 and 400~kPa ($S_P=0$ indicates no change, $S_P=1$ represents a twofold increase, and $S_P=-1$ represents a twofold decrease).}
\label{fig:flat_pressureandfig:pressure_log}
\end{figure}

The flat compliant surface provides the simplest geometry for evaluating the coupled effects of nozzle diameter and supply pressure. Figure~\ref{fig:flat_pressure} presents the mean peak torque as a function of supply pressure. The numerical results are summarized in Table~\ref{tab:flat_results}.

\begin{table}[t]
\centering
\scriptsize
\caption{Torque for the Flat Compliant Surface [N$\cdot$mm]}
\label{tab:flat_results}
\setlength{\tabcolsep}{2.7pt}
\renewcommand{\arraystretch}{1.15}
\begin{tabular}{|c|c|c|c|c|}
\hline
\textbf{$d_n$} &
\textbf{100 kPa} &
\textbf{200 kPa} &
\textbf{300 kPa} &
\textbf{400 kPa} \\
\hline
0.6 mm &
$0.238\pm0.048$ &
$0.212\pm0.022$ &
$0.307\pm0.110$ &
$0.282\pm0.064$ \\
\hline
0.8 mm &
$0.620\pm0.050$ &
$0.757\pm0.080$ &
$0.952\pm0.061$ &
$0.963\pm0.110$ \\
\hline
1.0 mm &
$0.541\pm0.096$ &
$0.715\pm0.218$ &
$0.994\pm0.267$ &
$1.601\pm0.376$ \\
\hline
\end{tabular}
\end{table}

With the 0.6 mm nozzles, the vortex-induced torque remained comparatively small and showed no clear monotonic dependence on pressure. The mean torque was $0.238\pm0.048$~N$\cdot$mm at 100~kPa and 0.282$\pm$0.064~N$\cdot$mm at 400~kPa, corresponding to an increase of only 18.6\%. A substantially stronger pressure dependence was obtained when the nozzle diameter was increased. For the G$_2$ ($d_n$=0.8 mm) configuration, the mean torque increased from $0.620\pm0.050$~N$\cdot$mm at 100~kPa to $0.963\pm0.110$~N$\cdot$mm at 400~kPa, representing an increase of 55.3\%. The strongest pressure dependence was obtained using the G$_3$ gripper. The corresponding torque increased from $0.541\pm0.096$~N$\cdot$mm to $1.601\pm0.376$~N$\cdot$mm, corresponding to an increase of approximately 196\%. The results indicate that increasing nozzle diameter does not simply increase torque by a constant factor. Instead, one of its most important effects is to increase the sensitivity of the generated torque to the supplied pressure.

The torque-dependent behavior observed for the flat surface was also reflected across the complete geometry set. Considering all 41 investigated surfaces, the median changes in mean peak vortex-induced torque between 100 and 400~kPa were 23.6\%, 53.7\%, 111.6\% for the G$_1$, G$_2$, and G$_3$ grippers, respectively. Torque increased from 100 to 400~kPa in 25 of 41 geometries for G$_1$, 34 of 41 geometries for G$_2$, and 38 of 41 geometries for G$_3$. The torque dependence is summarized using the logarithmic metric of Eq.~\eqref{eq:pressure_log_metric} in Fig.~\ref{fig:pressure_log}. The logarithmic representation was selected because it preserves all experimental conditions while reducing visual domination by very large percentage changes.

Table~\ref{tab:pressure_geometry} summarizes the median torque response of the four curved surface families. The nozzle-dependent effect is particularly clear for the cylindrical concave surfaces. Their median response changes from a slight reduction of 9.5\% with the 0.6 mm nozzles to an increase of 113.7\% with the 1.0 mm nozzles.

\begin{table}[t]
\centering
\scriptsize
\caption{Median Change in Mean Peak Torque Between 100 and 400 kPa}
\label{tab:pressure_geometry}
\setlength{\tabcolsep}{3.5pt}
\renewcommand{\arraystretch}{1.15}
\begin{tabular}{|l|c|c|c|c|c|c|}
\hline
\multirow{2}{*}{\textbf{Geometry}} &
\multicolumn{2}{c|}{\textbf{0.6 mm}} &
\multicolumn{2}{c|}{\textbf{0.8 mm}} &
\multicolumn{2}{c|}{\textbf{1.0 mm}} \\
\cline{2-7}
&
\textbf{$\Delta M_z$} &
\textbf{Positive} &
\textbf{$\Delta M_z$} &
\textbf{Positive} &
\textbf{$\Delta M_z$} &
\textbf{Positive} \\
\hline
Cylindrical concave &
$-9.5\%$ & 3/10 &
$35.8\%$ & 7/10 &
$113.7\%$ & 10/10 \\
\hline
Cylindrical convex &
$50.8\%$ & 8/10 &
$39.4\%$ & 7/10 &
$109.4\%$ & 10/10 \\
\hline
Dome concave &
$19.5\%$ & 6/10 &
$57.6\%$ & 9/10 &
$63.2\%$ & 7/10 \\
\hline
Dome convex &
$91.5\%$ & 7/10 &
$65.7\%$ & 10/10 &
$140.0\%$ & 10/10 \\
\hline
\end{tabular}
\end{table}

The results demonstrate that supply pressure, nozzle diameter, and surface geometry are strongly coupled. A larger nozzle increases the available tangential inlet area and can increase the airflow introduced into the chamber as pressure rises. However, the resulting torque additionally depends on how efficiently the rotating airflow remains coupled to the surface outlet through the gap.

\subsection{Surface Influence}
\label{subsec:cylindrical_results}

\begin{figure}[t]
\centering
\begin{subfigure}[t]{0.32\linewidth}
\centering
\includegraphics[width=\linewidth]{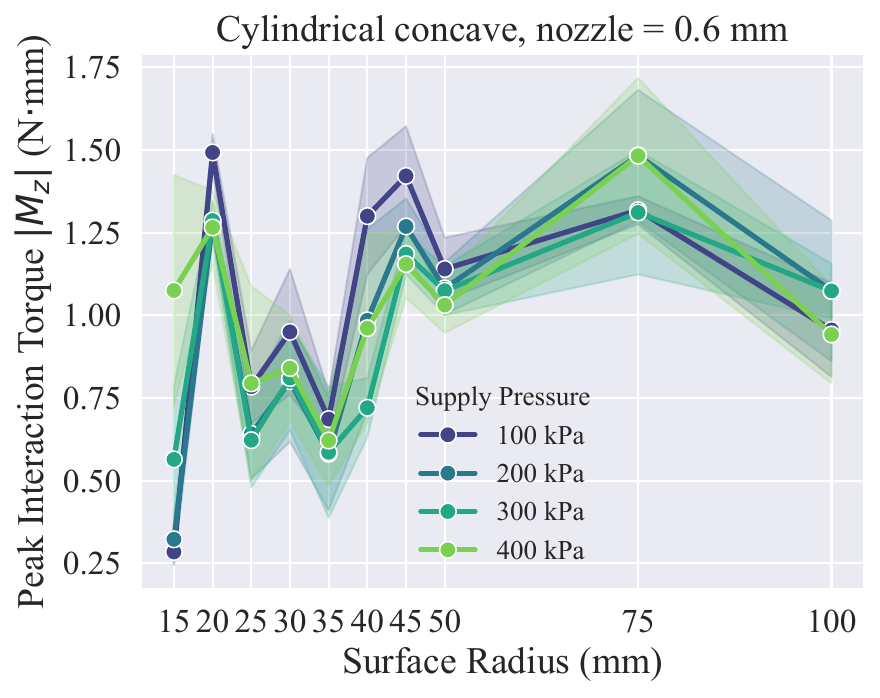}
\vspace{-5.5mm}
\caption{$d_n=0.6$ mm}
\end{subfigure}
\hfill
\begin{subfigure}[t]{0.32\linewidth}
\centering
\includegraphics[width=\linewidth]{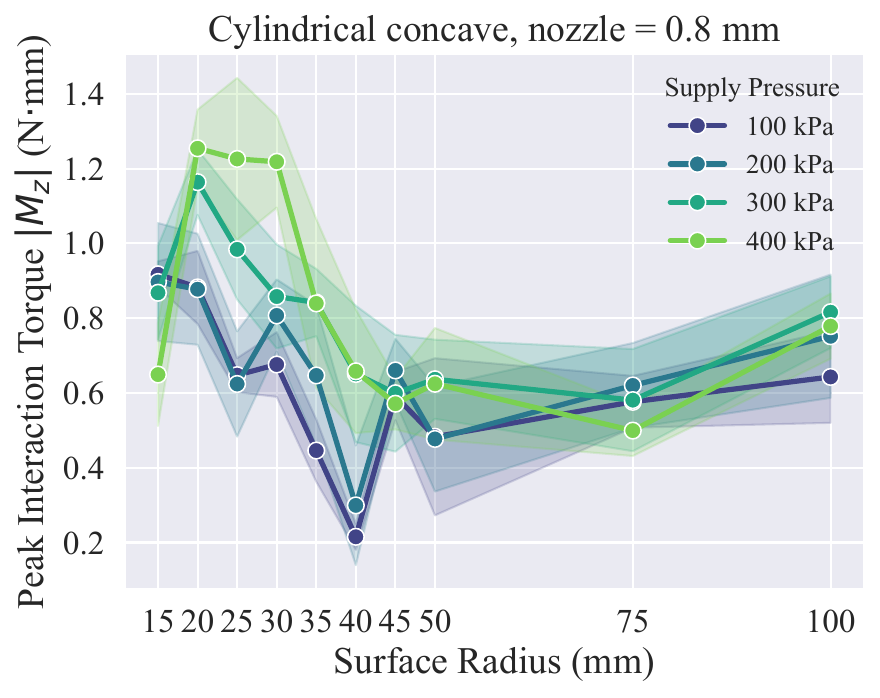}
\vspace{-5.5mm}
\caption{$d_n=0.8$ mm}
\end{subfigure}
\hfill
\begin{subfigure}[t]{0.32\linewidth}
\centering
\includegraphics[width=\linewidth]{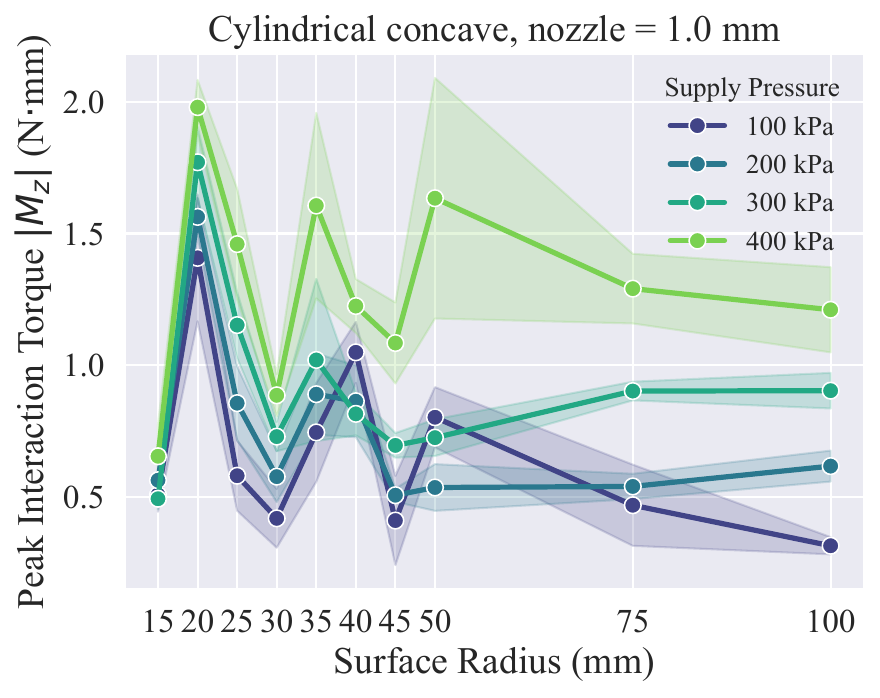}
\vspace{-5.5mm}
\caption{$d_n=1.0$ mm}
\end{subfigure}\\
\begin{subfigure}[t]{0.32\linewidth}
\centering
\vspace{1mm}
\includegraphics[width=\linewidth]{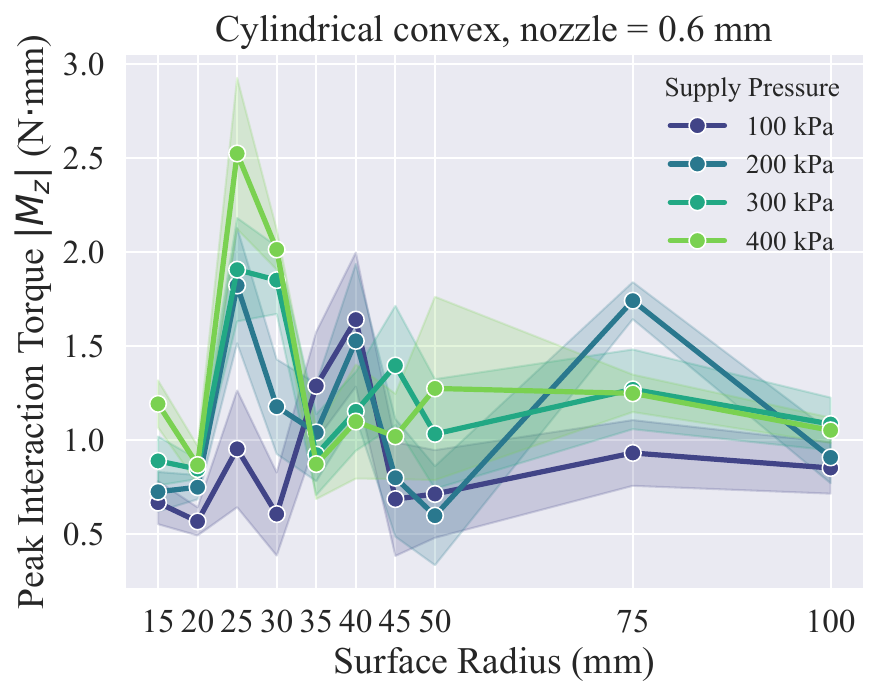}
\vspace{-5.5mm}
\caption{$d_n=0.6$ mm}
\end{subfigure}
\hfill
\begin{subfigure}[t]{0.32\linewidth}
\centering
\vspace{1mm}
\includegraphics[width=\linewidth]{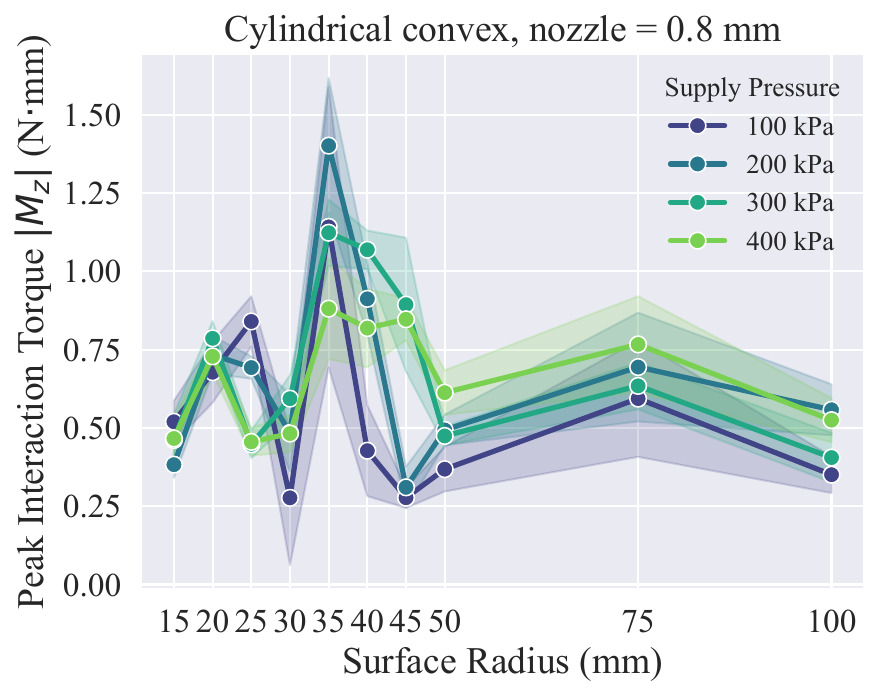}
\vspace{-5.5mm}
\caption{$d_n=0.8$ mm}
\end{subfigure}
\hfill
\begin{subfigure}[t]{0.32\linewidth}
\centering
\vspace{1mm}
\includegraphics[width=\linewidth]{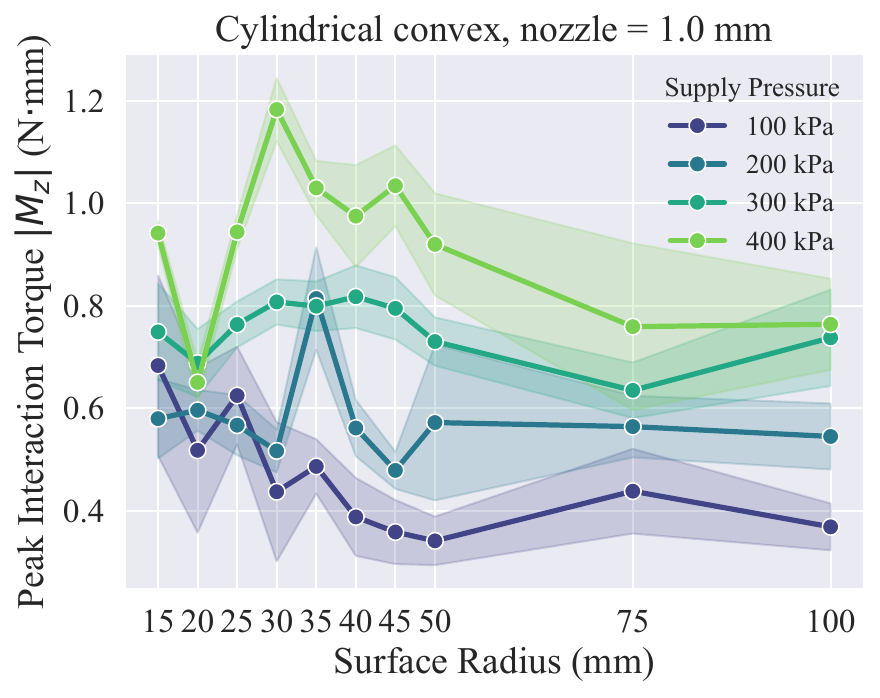}
\vspace{-5.5mm}
\caption{$d_n=1.0$ mm}
\end{subfigure}
\vspace{-1.5mm}
\caption{Mean peak torque as a function of radius for cylindrical surfaces at different supply pressures: (a, b, c) - concave; (d, e, f) - convex.}
\label{fig:cyl_concave}
\end{figure}

The radius-dependent torque for the cylindrical concave and convex surfaces can be seen in Figure~\ref{fig:cyl_concave}. The cylindrical surfaces do not show a universal monotonic dependence of torque on radius. Furthermore, the difference between corresponding concave and convex cylindrical surfaces is comparatively small when all matched experimental conditions are considered.

The median concave-to-convex torque ratio is 1.06, which is close to unity and indicates that the reversal of cylindrical curvature alone does not consistently produce a large change in torque. The reason is that a cylindrical surface is curved along only one principal direction. The perpendicular direction remains locally flat, leaving portions of the peripheral airflow path comparatively unaffected by the cylinder radius. As a result, cylindrical curvature modifies the flow distribution locally without producing strong circumferential confinement around the complete gripper.

\begin{figure}[t]
\centering
\begin{subfigure}[t]{0.32\linewidth}
    \centering
    \includegraphics[width=\linewidth]{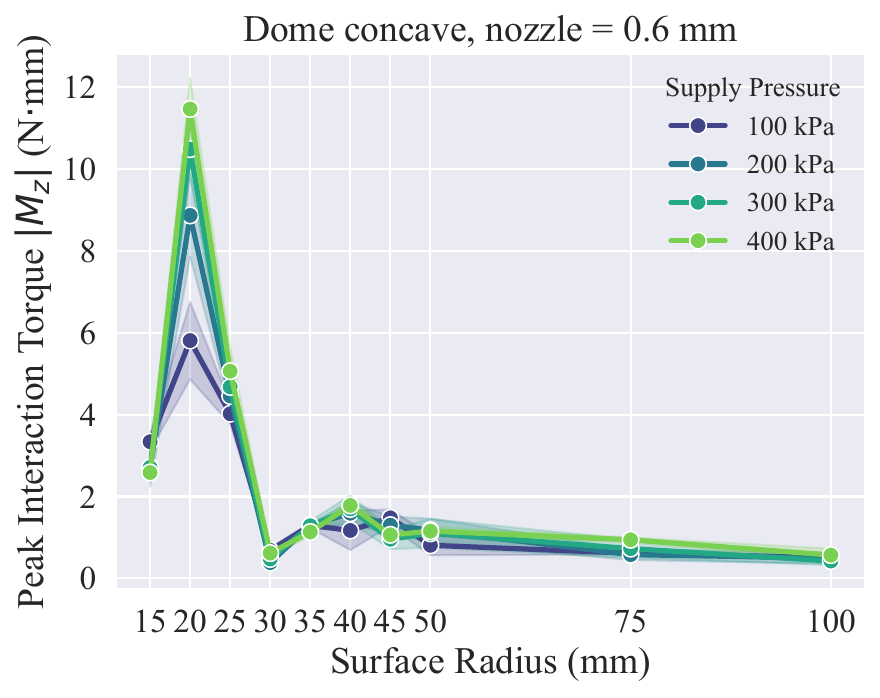}
    \vspace{-5.5mm}
    \caption{$d_n=0.6$ mm}
    \end{subfigure}
    \hfill
    \begin{subfigure}[t]{0.32\linewidth}
    \centering
    \includegraphics[width=\linewidth]{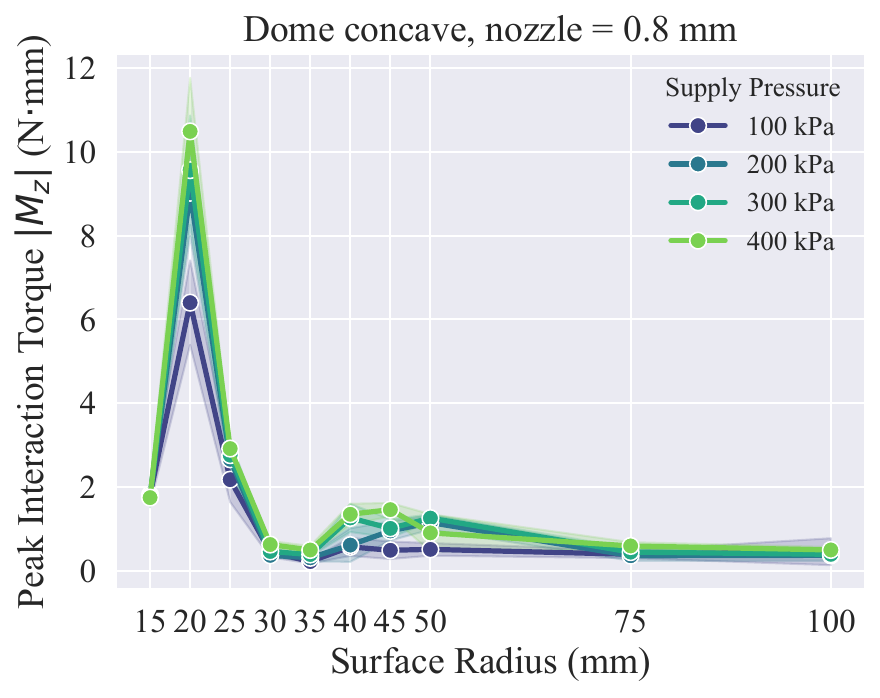}
    \vspace{-5.5mm}
    \caption{$d_n=0.8$ mm}
    \end{subfigure}
    \hfill
    \begin{subfigure}[t]{0.32\linewidth}
    \centering
    \includegraphics[width=\linewidth]{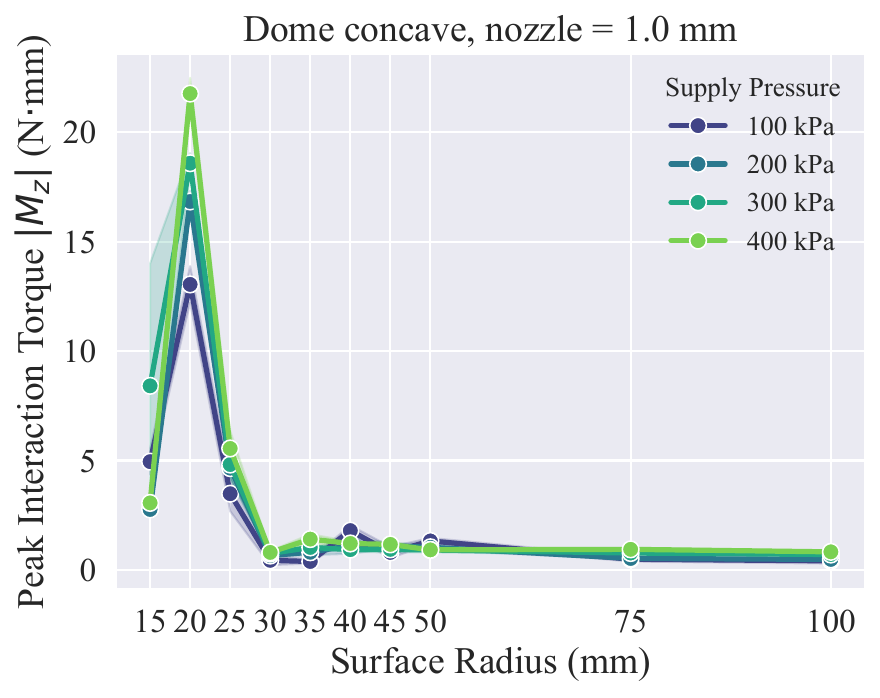}
    \vspace{-5.5mm}
    \caption{$d_n=1.0$ mm}
    \end{subfigure}\\
    \begin{subfigure}[t]{0.32\linewidth}
    \centering
    \vspace{1mm}
    \includegraphics[width=\linewidth]{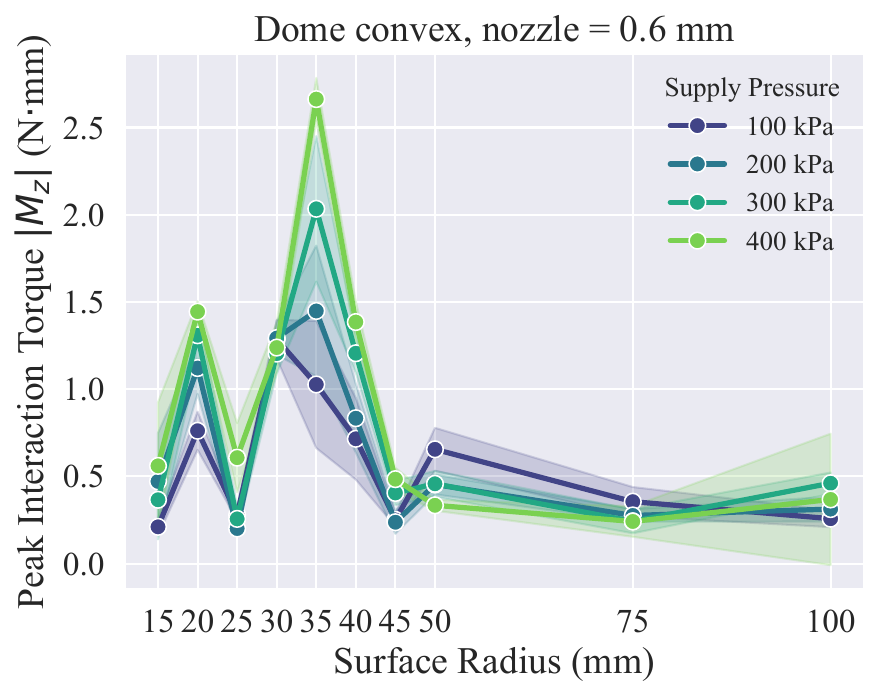}
    \vspace{-5.5mm}
    \caption{$d_n=0.6$ mm}
    \end{subfigure}
    \hfill
    \begin{subfigure}[t]{0.32\linewidth}
    \centering
    \vspace{1mm}
    \includegraphics[width=\linewidth]{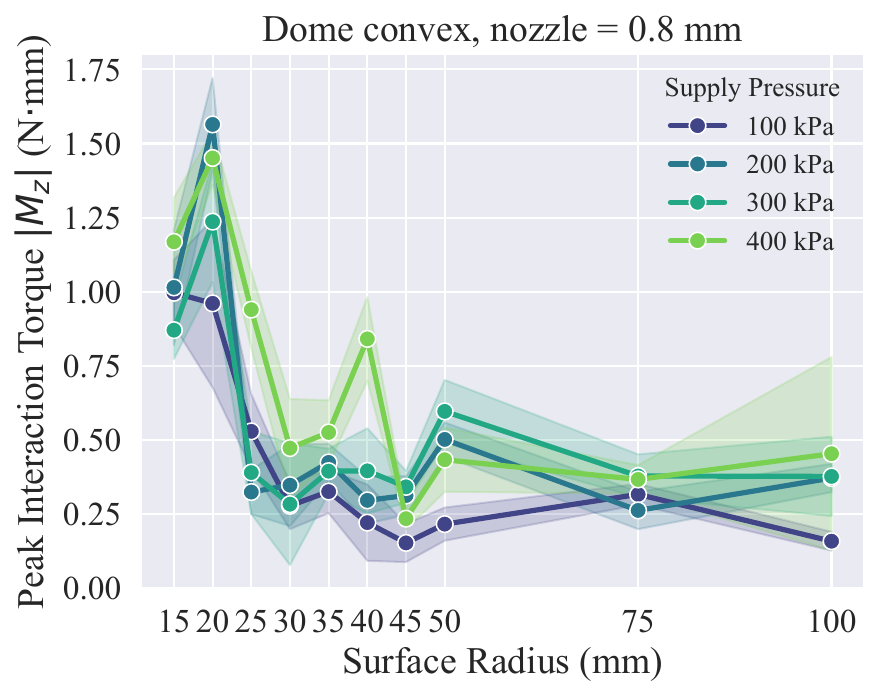}
    \vspace{-5.5mm}
    \caption{$d_n=0.8$ mm}
    \end{subfigure}
    \hfill
    \begin{subfigure}[t]{0.32\linewidth}
    \centering
    \vspace{1mm}
    \includegraphics[width=\linewidth]{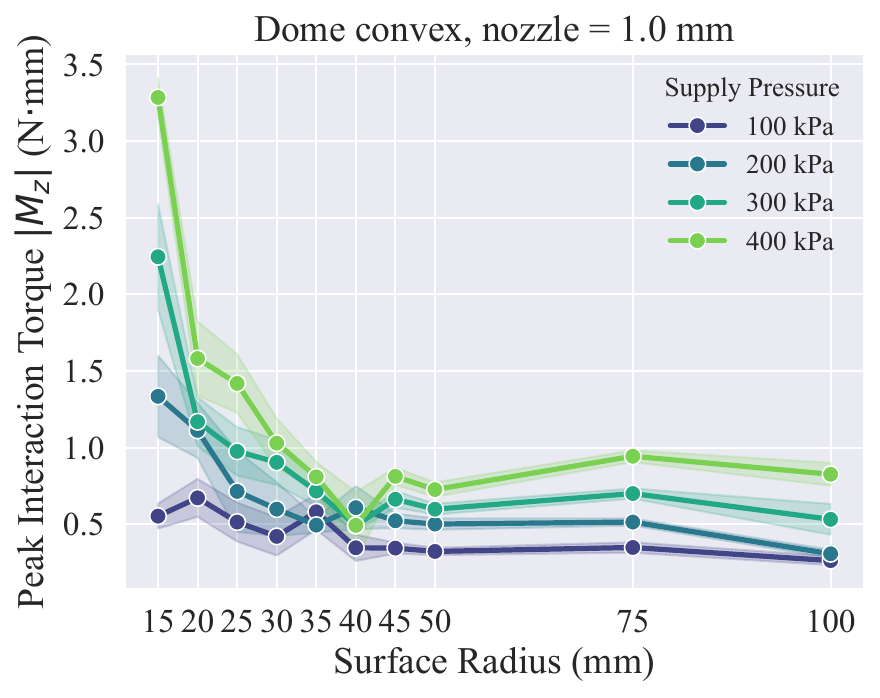}
    \vspace{-5.5mm}
    \caption{$d_n=1.0$ mm}
    \end{subfigure}
\vspace{-1.5mm}
\caption{Mean peak torque as a function of radius for dome surfaces at different supply pressures: (a, b, c) - concave; (d, e, f) - convex.}
\label{fig:dome_concave}
\end{figure}

The dome geometries show a substantially stronger radius dependence (Fig.~\ref{fig:dome_concave}). For dome-convex surfaces, smaller radii generally produce larger torque, particularly for the G$_3$ gripper configuration. With the 1.0 mm nozzles, the Spearman correlation between radius and torque is negative at every pressure $\rho_s = -0.830, -0.842, -0.842, -0.648$ for 100, 200, 300, and 400~kPa, respectively. The corresponding relationships are statistically significant at all four pressure levels ($p<0.05$). For example, using G$_3$ at 400~kPa, the $R=15$~mm dome-convex surface produces $\overline{M}_{z,\mathrm{peak}} =3.285$~N$\cdot$mm, whereas the $R=100$~mm surface produces only 0.826~N$\cdot$mm. 

The dome-concave response is even more distinctive because it contains a local maximum at $R=20$~mm. Consequently, the measured torque cannot be represented using a simple rule such as increasing torque with decreasing radius. Instead, the data indicate a geometry-specific interaction between surface curvature and the characteristic dimensions of the vortex gripper. 

\begin{figure}[t]
\centering
\begin{subfigure}[t]{0.48\linewidth}
\centering
\includegraphics[width=\linewidth]{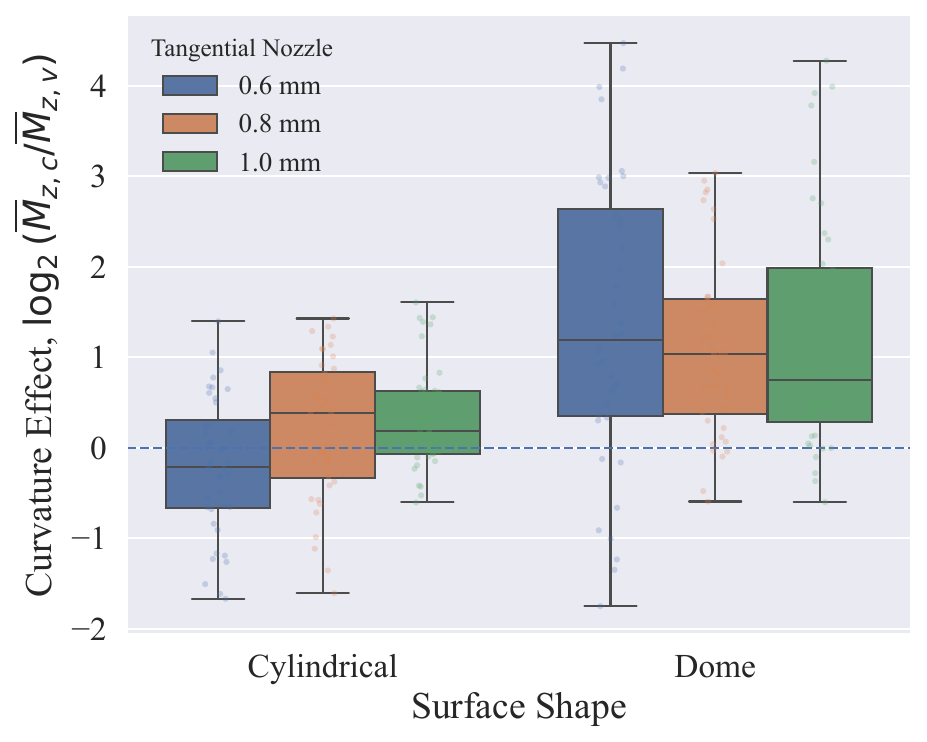}
\vspace{-5.5mm}
\caption{}
\label{fig:concave_convex_log}
\end{subfigure}
\hfill
\begin{subfigure}[t]{0.48\linewidth}
\centering
\includegraphics[width=\linewidth]{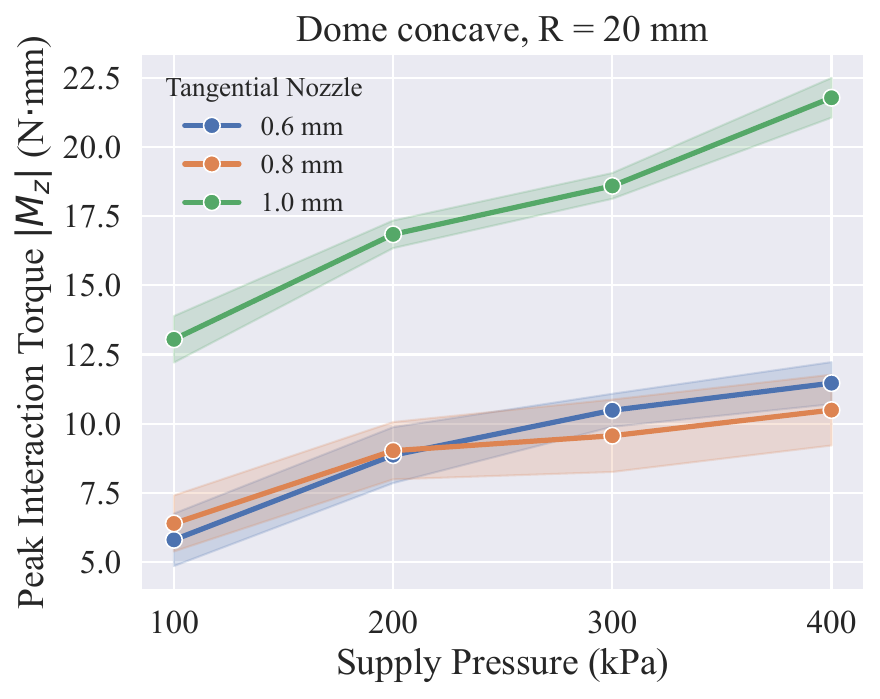}
\vspace{-5.5mm}
\caption{}
\label{fig:r20_pressure}
\end{subfigure}
\vspace{-2mm}
\caption{(a) Logarithmic concave-to-convex torque ratio for matched surface radius, pressure, and nozzle diameter. (b) Mean peak torque for the $R=20$~mm dome-concave surface as a function of supply pressure.}
\label{fig:concave_convex_logandfig:r20_pressure}
\end{figure}

The logarithmic ratio \eqref{eq:concave_convex_metric} compares the corresponding concave and convex surfaces present in Fig.~\ref{fig:concave_convex_log}. For cylindrical surfaces, the median raw ratio is only 1.06; for dome geometries, it is 1.99. Thus, the dome-concave surfaces generate approximately twice the torque of the corresponding dome-convex surfaces in the median matched condition. This effect is not produced solely by the exceptionally large $R=20$~mm results. When all $R=20$~mm pairs are excluded, the median dome ratio remains 1.74. The observed asymmetry, therefore, indicates a general influence of the three-dimensional orientation. For a cylindrical surface, one principal surface direction remains uncurved. In contrast, for a dome, the curvature modifies the gripper-surface gap around the complete circumference. A concave dome may, therefore, restrict the peripheral leakage path in all radial directions, whereas a convex dome tends to increase the peripheral gap.

The most clearly geometry-dependent response in the complete dataset occurs for the dome-concave surface with $R=20$~mm (Fig.~\ref{fig:r20_pressure}). At every investigated pressure and nozzle diameter, the $R=20$~mm dome-concave geometry produces substantially greater torque than most of the remaining dome-concave surfaces. Relative to the next-highest radius under the same operating condition, the amplification ranges from approximately 1.44 to 3.92. Relative to the median of the remaining dome-concave radii, the amplification ranges from approximately 4.97 to 20.09. The maximum mean torque measured in the complete experiment is $\overline{M}_{z,\mathrm{peak}}=21.770\pm0.717$~N$\cdot$mm, obtained using G$_3$ at 400~kPa. The corresponding coefficient of variation is only $CV=3.29$\%. The large response is therefore highly repeatable and cannot reasonably be attributed to a small number of abnormal trials.

Based on our previous vortex-gripper studies and CFD analysis~\cite{mykhailyshyn2026icra,roman2026investigating}, we interpret the pronounced response at $R=20$~mm in terms of gap formation between the compliant object and the outer gripper surface, whose diameter is $D_g=20$~mm, rather than the inner vortex-chamber surface. This peripheral clearance governs the airflow discharge and may modify vortex confinement and tangential momentum transfer, providing a physical explanation for the strong geometry dependence of the measured torque. However, the specific torque maximum at $R=20$~mm has not been directly verified by flow-field measurements or CFD for the present configuration.

\subsection{ML Prediction of Vortex-Induced Torque}
\label{subsec:ml_results}



Among the five evaluated regression approaches, GPR achieved the
highest mean fold-level performance under repeated 5-fold
cross-validation $R^2=0.899\pm0.080$, $RMSE=0.508\pm0.217$~N$\cdot$mm, with an MAE of $0.240\pm0.043$~N$\cdot$mm. Extra Trees and Random Forest followed, with mean $R^2$ values of $0.885\pm0.101$ and $0.870\pm0.130$. The Random Forest + AdaBoost ensemble and AdaBoost produced lower mean $R^2$ values of 0.803 and 0.668. GPR was therefore selected as the final predictive model.

\begin{figure}[t]
\centering
\includegraphics[width=0.72\linewidth]{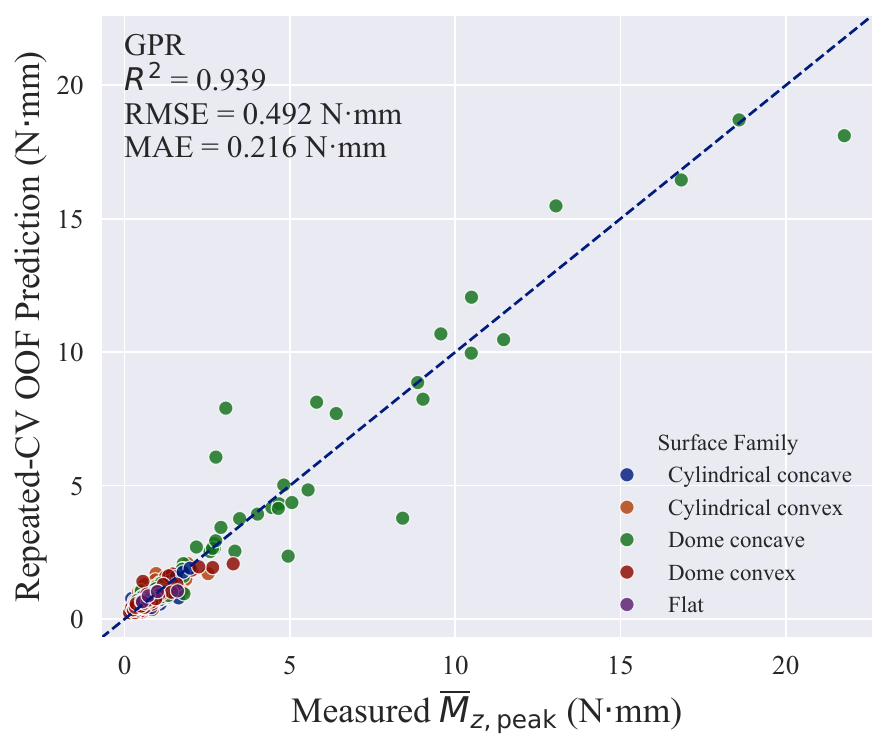}
\vspace{-2mm}
\caption{Measured and repetition-averaged out-of-fold predictions of
peak vortex-induced torque for the selected GPR model.}
\label{fig:ml_oof}
\end{figure}

For the condition-level comparison shown in Fig.~\ref{fig:ml_oof}, the five repeated out-of-fold predictions obtained for each condition were averaged. The resulting aggregated predictions yielded $R^2=0.939$, $RMSE=0.492$~N$\cdot$mm, $MAE=0.216$~N$\cdot$mm. The concentration of the predictions around the ideal $1{:}1$ line demonstrates that the combined pneumatic and geometry-based descriptors provide sufficient information to predict the peak torque within the experimentally sampled design domain.

\begin{figure}[t]
\centering
\vspace{-4mm}
\includegraphics[width=0.82\linewidth]{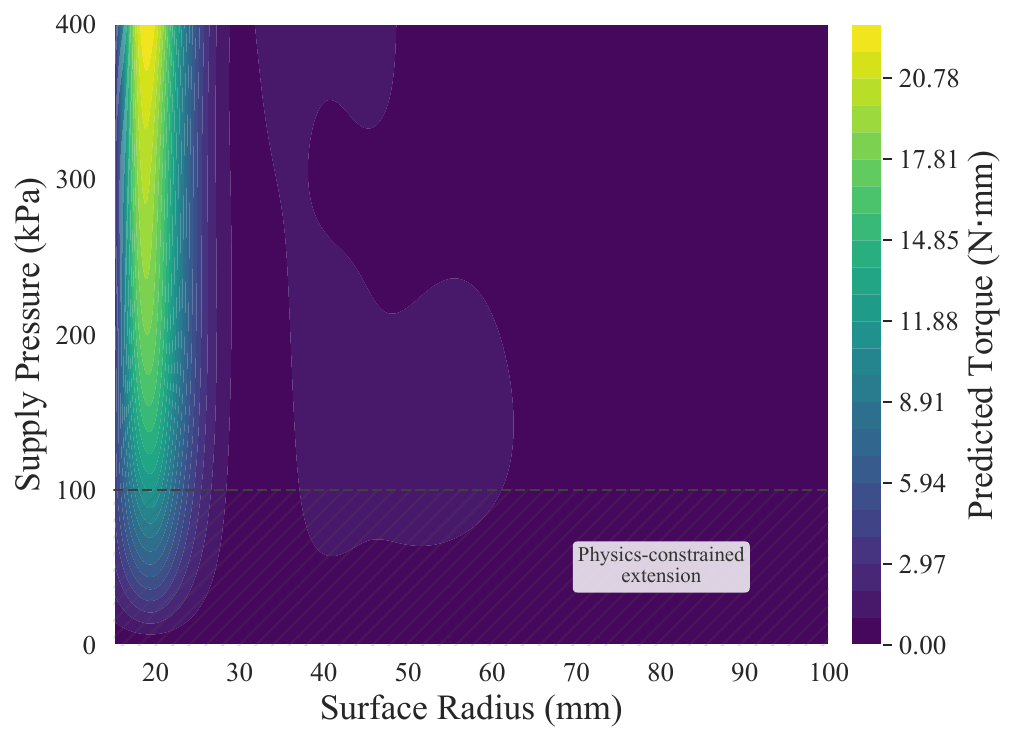}
\vspace{-2mm}
\caption{GPR prediction torque map for the vortex gripper (G$_3$) with dome-concave surface.}
\label{fig:ml_torque_map}
\end{figure}

The combined influence of surface radius and supply pressure was subsequently represented using the predicted torque map in Fig.~\ref{fig:ml_torque_map}. The map was generated over the experimentally supported pressure interval of 100-400~kPa and the radius interval of 0-100~mm.

The torque map clearly identifies a localized high-torque region near $R\approx19$-20~mm, with its magnitude increasing strongly with supply pressure. Away from this region, the predicted torque decreases substantially and varies more gradually with radius. The resulting map, therefore, provides a compact representation of the strongly coupled effects of surface geometry and pneumatic operating conditions.

A physical boundary condition $\widehat{M}_{z,\mathrm{peak}}=0$ at $P_s=0$ was imposed in the prediction procedure. Predictions in the unmeasured pressure range $0<P_s<100$~kPa were generated using the learned relationship between pneumatic and geometric parameters, together with this zero-pressure constraint.

Overall, the machine-learning analysis demonstrates that vortex-induced torque can be represented by a continuous geometry/pressure-dependent model within the investigated design domain. The model predictions reproduce the geometry-dependent trends in the experimental dataset but do not independently validate the proposed flow mechanism. The machine-learning analysis, therefore, indicates that surface geometry is not only responsible for the large differences observed experimentally but also provides the principal information required for predicting the resulting rotational interaction.

\section{Conclusion}
\noindent This study investigated vortex-induced torque during the grasping of compliant objects and quantified the coupled effects of supply pressure, tangential nozzle diameter, surface geometry, and curvature. The results show that increasing the tangential nozzle diameter primarily increases the sensitivity of the generated torque to supply pressure rather than producing a uniformly higher torque. Across all investigated geometries, the median increase in peak torque between 100 and 400~kPa was 23.6\%, 53.7\%, and 111.6\% for nozzle diameters of 0.6, 0.8, and 1.0~mm, respectively. Surface geometry had an equally important influence on the rotational interaction. Dome-concave surfaces generated approximately twice the torque of corresponding dome-convex surfaces in the median matched condition, whereas the difference between cylindrical concave and convex surfaces was substantially smaller. The strongest response occurred for the $R=20$~mm dome-concave geometry, which reached $21.770\pm0.717$~N$\cdot$mm, with the 1.0 mm nozzles at 400~kPa. A comparison of five machine-learning regression models showed that Gaussian Process Regression provided the best overall predictive performance, achieving $R^2=0.899\pm0.080$ under repeated 5-fold cross-validation. The high magnitude and low experimental variation indicate that this response represents a repeatable geometry-dependent phenomenon rather than an isolated extreme value.

Overall, the results demonstrate that vortex-gripper performance cannot be characterized solely by lifting force or supply pressure. This effect should be used for the dexterous rotation of objects being manipulated in both manufacturing and medical applications. Future work will focus on the dexterous manipulation of objects using torque, combining the vortex-induced torque effect.



\bibliographystyle{IEEEtran}
\IEEEtriggercmd{\enlargethispage{2in}}
\bibliography{Dex}


\end{document}